\documentclass[11pt]{article}

\usepackage[preprint]{acl}

\usepackage{times}
\usepackage{latexsym}
\usepackage{amsmath}
\usepackage{amssymb}

\usepackage[T1]{fontenc}
\usepackage[table]{xcolor}
\usepackage[utf8]{inputenc}
\usepackage{tcolorbox}
\tcbuselibrary{most}
\usepackage{microtype}

\usepackage{inconsolata}
\usepackage{array}
\usepackage{booktabs}
\usepackage{placeins}
\usepackage{graphicx}
\newlength\savewidth\newcommand\shline{\noalign{\global\savewidth\arrayrulewidth
  \global\arrayrulewidth 1pt}\hline\noalign{\global\arrayrulewidth\savewidth}}
\definecolor{deepred}{HTML}{940000}
\definecolor{Gray}{gray}{0.94}
\hypersetup{linkcolor=deepred}
\hypersetup{urlcolor  = [rgb]{0.4,0.15,0.95}}
\hypersetup{citecolor=[rgb]{0.4,0.15,0.95}}
\usepackage{cleveref}
\definecolor{Gray}{gray}{0.94}

\newtcolorbox{rqbox}{
  enhanced,
  colback=gray!4,
  colframe=black!70,
  boxrule=0.6pt,
  arc=2pt,
  left=6pt,
  right=6pt,
  top=6pt,
  bottom=6pt,
  borderline west={2pt}{0pt}{black!85},
  fonttitle=\bfseries,
  title={Research Question},
  attach boxed title to top left={xshift=6pt,yshift=-2mm},
  boxed title style={
    colback=black!85,
    colframe=black!85,
    arc=1pt,
    boxrule=0pt
  },
  coltitle=white,
  before skip=6pt,
  after skip=6pt
}
\newtcolorbox{ptbox}{
  enhanced,
  breakable,
  colback=gray!4,
  colframe=black!70,
  boxrule=0.6pt,
  arc=2pt,
  left=6pt,
  right=6pt,
  top=6pt,
  bottom=6pt,
  borderline west={2pt}{0pt}{black!85},
  fonttitle=\bfseries,
  title={Prompt Template},
  attach boxed title to top left={xshift=6pt,yshift=-2mm},
  boxed title style={
    colback=black!85,
    colframe=black!85,
    arc=1pt,
    boxrule=0pt
  },
  coltitle=white,
  before skip=6pt,
  after skip=6pt
}

\newcommand{\physhack}{\textit{PhysHack}}

\title{Sample-Efficient Post-Training for LEGO Spatial-Physics Reasoning}

\author{%
  \begin{tabular}{c}
    \textbf{Yuhuan Yuan}$^{1,\dagger}$\thanks{$^{\dagger}$ Equal contribution.
Correspondence to \href{mailto:yyuan065@connect.hkust-gz.edu.cn}{Yuhuan Yuan}. } \quad
    \textbf{Zhouliang Yu}$^{2,\dagger}$ \quad
    \textbf{Minghao Liu}$^{3}$ \quad
    \textbf{Weiyang Liu}$^{2}$ \quad
    \textbf{Ge Lin Kan}$^{1}$ \\
    {\normalfont $^{1}$HKUST(GZ) \quad
    $^{2}$CUHK} \quad 
    {\normalfont $^{3}$ZODA} \\
    {\normalfont \href{https://yuhuanyuan.github.io/lego_rl}{hkust-gz.spatial.ai}}
  \end{tabular}%
}

\begin{document}
\raggedbottom
\setlength{\abovedisplayskip}{2pt}
\setlength{\belowdisplayskip}{2pt}
\setlength{\abovedisplayshortskip}{2pt}
\setlength{\belowdisplayshortskip}{2pt}
\maketitle
\begin{abstract}
LLM-based LEGO assembly requires both semantic grounding and physical feasibility.
In this paper, we identify a data-induced failure mode, \physhack{}, in which generated assemblies satisfy physical-validity constraints while remaining geometrically misaligned, semantically inconsistent, or poorly calibrated.
To address this challenge, we propose a model-based data selection approach that uses only $5\%$ of the original training data while improving semantic and structural alignment across multiple independent evaluators.
We further introduce PVPO, a reinforcement learning method that couples physical-validity and voxel-space geometric rewards to strengthen the LEGO synthesis capabilities of LLMs.
PVPO-trained DeepSeek-Llama-8B outperforms the frontier VLM Kimi-K3 across semantic, structural, and physical evaluation dimensions, even when Kimi-K3 is provided with a ground-truth image as an additional reference.

\end{abstract}

\begin{figure}[!t]
    \centering
    \includegraphics[width=0.95\columnwidth]{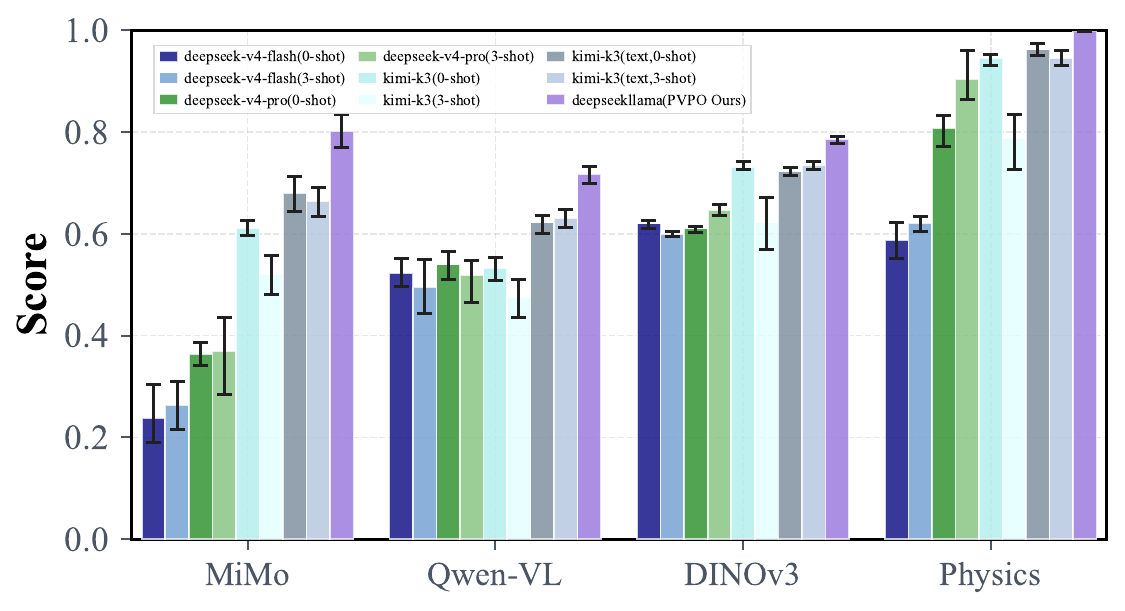}
    \vspace{0.6em}

    \includegraphics[width=0.95\columnwidth]{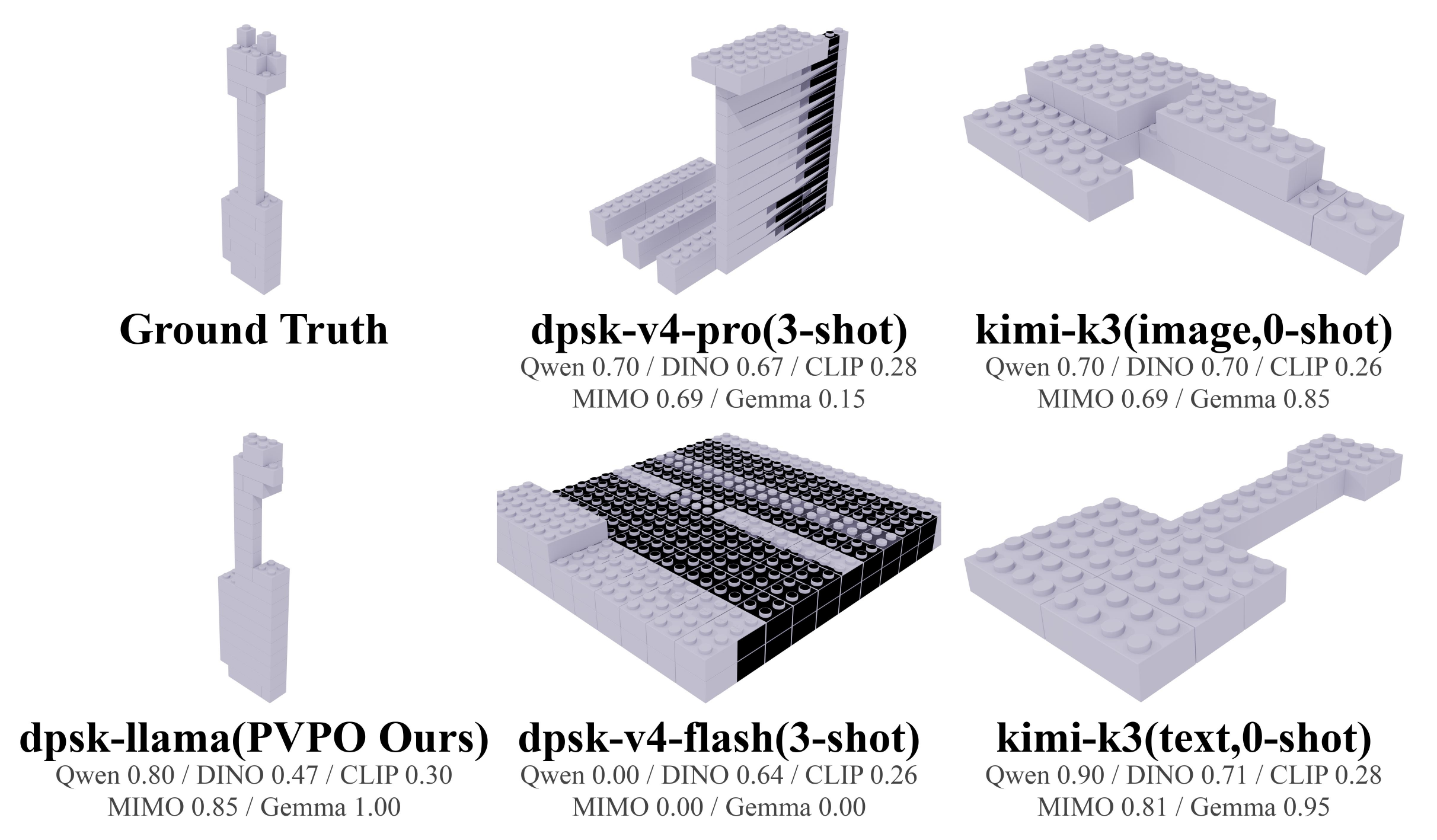}
    \caption{\textbf{Quantitative and qualitative comparison of DeepSeek-V4, Kimi-K3, and PVPO-trained DeepSeek-Llama-8B on LBA.} \textbf{Top:} Performance across semantic and physical evaluation metrics; error bars show statistical variability across evaluation runs. \textbf{Bottom:} Qualitative comparison on the ``Guitar'' synthesis task, black bricks indicate collisions.}
    \label{fig:frontier-grouped-error-bars}
    \label{fig:case10-frontier}
\end{figure}

\begin{figure*}[t]
    \centering
    \includegraphics[width=0.80\textwidth]{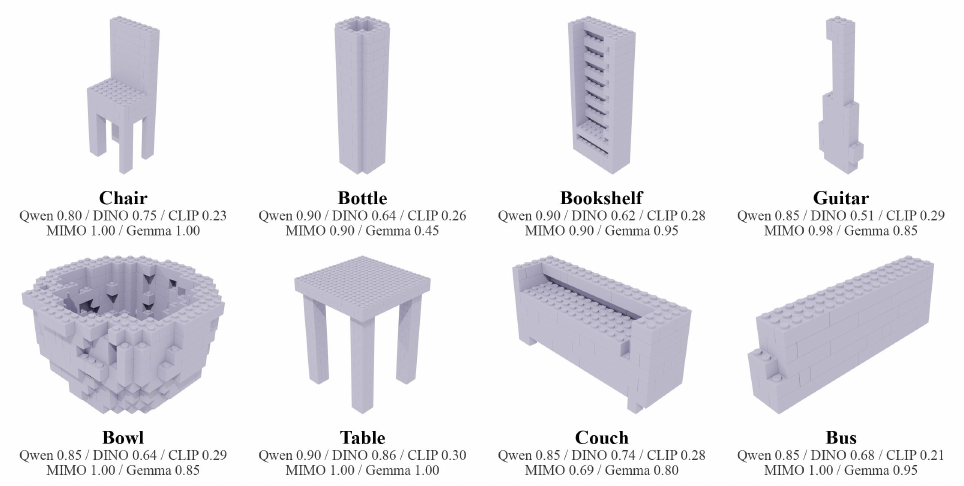}
    \caption{\textbf{Qualitative examples of PVPO-generated LEGO structures across diverse object categories.} The scores shown below each example correspond to MiMo, Gemma, Qwen-VL, DINOv3, and CLIP evaluations.}
    \label{fig:pvpo-cases}
\end{figure*}

\section{Introduction}
LEGO Brick Assembly (LBA) \citep{pun2025generating,kulits2026bricknet,ahn2022budget} creates real-world objects from modular LEGO bricks, challenging the creativity as well as the precise spatial and physical reasoning abilities of generative models \citep{kingma2013auto,ho2020denoising,vaswani2017attention}. 
Recent advances in large language models (LLMs) \citep{grattafiori2024llama,qwen2024qwen2,achiam2023gpt} reframed LBA into a program synthesis task, where models generate executable assembly programs through language modeling.
However, how post-training data shapes the LBA capabilities of LLMs remains underexplored. 
Although existing LBA datasets contain more than 200,000 examples \citep{kulits2026bricknet,pun2025generating}, they contain substantial noise, redundancy, and annotation errors, making it difficult to understand the role of data in LBA post-training (See \Cref{fig:origin_image}). Performance may depend not only on data scale, but also on which examples provide useful supervision for spatial-physics reasoning. 
Our failure-case analysis suggests that models trained on the full-scale dataset from \citet{pun2025generating} may exhibit unexpected behaviors, such as satisfying physical constraints but failing to preserve the intended object semantics. 
This motivates us to ask:
\emph{What makes a training example valuable for LEGO spatial-physics reasoning, and can a compact set of high-value examples enable more effective post-training than full-scale noisy supervision?}

\begin{figure*}[t]
    \centering
    \includegraphics[width=0.90\textwidth]{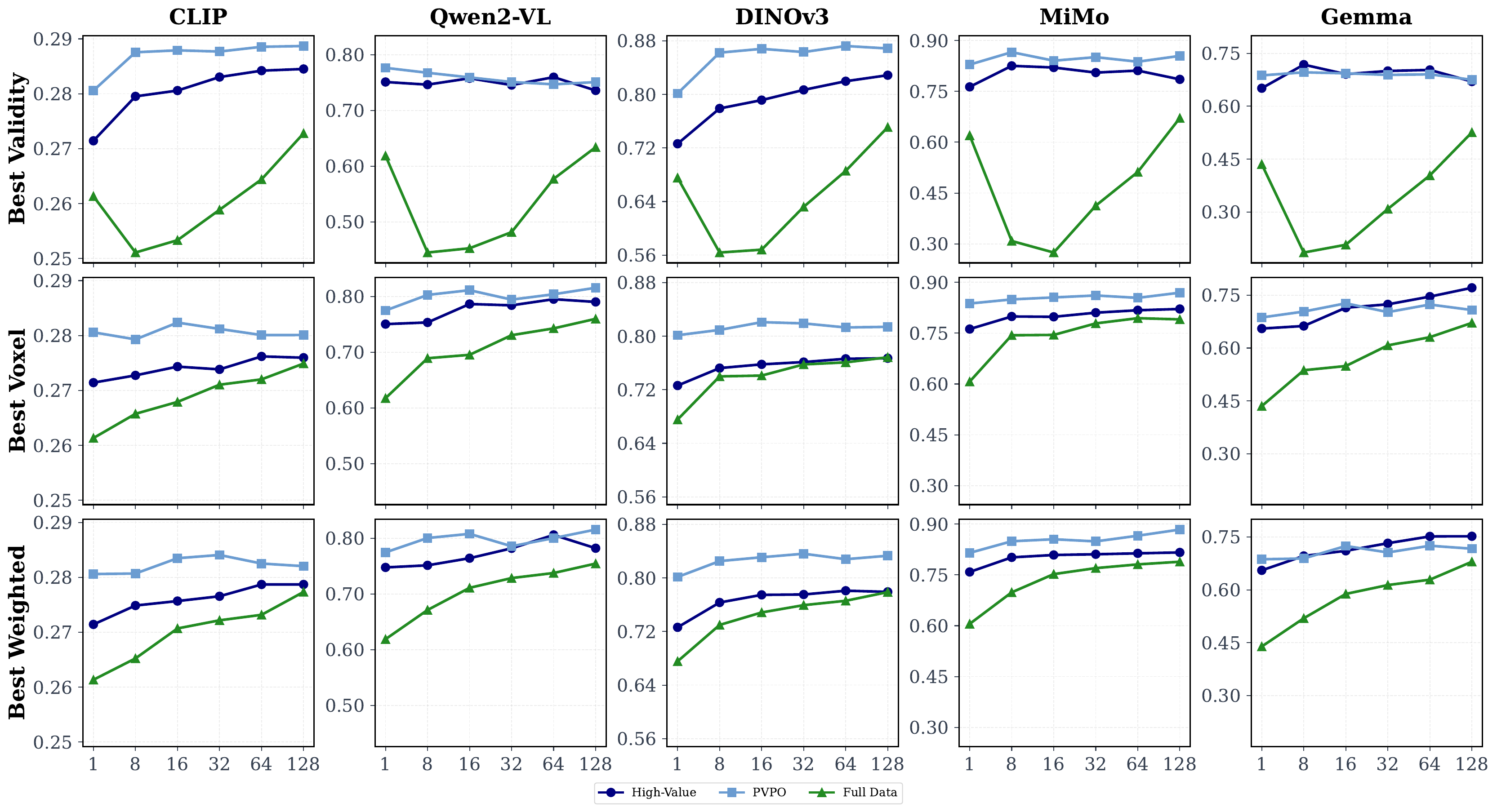}
    \caption{\textbf{Test-Time Scaling on Physics--Structure Alignment.} Best@K results evaluated by Qwen2-VL, CLIP, MiMo, Gemma, and DINOv3 under different selection metrics. PVPO consistently outperforms the full-dataset training baseline.}
    \label{fig:passk-vision}
\end{figure*}

We first study LBA post-training through the lens of data valuation \citep{koh2017understanding, jia2019towards, he2016dual,du2024chinese,liu2024makes}.
Given a large pool of noisy LEGO reasoning trajectories, we ask which examples provide effective supervision for learning spatial-physics reasoning.
Rather than treating all training examples as equally useful, we estimate their value by measuring the semantic consistency between the text description and the rendered LEGO structure, while filtering out examples that violate basic physical constraints.
By this, we construct a compact but high-value training subset, and to isolate the role of data quality and composition from that of data scale.
This further motivates a second question:
\emph{If valuable data can improve post-training efficiency, how should the learning objective better exploit the verifiable spatial and physical structure of LEGO assemblies?}
Data selection identifies useful supervision, but it does not by itself specify how a model should balance geometrical scene alignment and physical validity during generation.
This is particularly important for LBA, where a generated program must satisfy low-level assembly constraints while still preserving the object semantics described by the prompt.
To address this question, we introduce PVPO, a physics-informed reinforcement learning framework for LEGO spatial-physics reasoning.
PVPO combines simulation-based physical-validity rewards with structure-aware geometric rewards, encouraging models to generate assemblies that are both physically feasible and semantically aligned with the intended 3D object structure.
By coupling data valuation with verifiable post-training feedback, our framework provides a sample-efficient approach to improving LLM-based LBA (See qualitative examples in \Cref{fig:pvpo-cases}).
Our key contributions are as follows:
\vspace{-2mm}
\begingroup
\setlength{\leftmargini}{1.2em}
\begin{itemize}
    \item \textbf{Data-Induced Physics-Semantics Misalignment.}
    We identify the \physhack{} phenomenon in LBA. Models trained on noisy full-scale trajectories achieve high physical validity, e.g., $0.93$ Validity@4 on Qwen, but remain weak in semantic alignment $0.68/0.39$ by MiMo@4 and Gemma@4 (See \Cref{tab:semantic_sft}).

    \item \textbf{Data Valuation for Sample-Efficient Post-Training.}
Motivated by \physhack{}, we systematically study what makes LBA trajectories valuable for post-training. By comparing semantic, diversity, perplexity, and length-based selection signals, we find that VLM-based semantic valuation, together with domain diversity, identifies the most effective supervision. With only the top $5\%$ examples, our method improves Gemma@4 from $0.39$ to $0.63$, MiMo@4 from $0.68$ to $0.84$, and DINOv3@4 from $0.67$ to $0.70$ for Qwen-2.5 (See \Cref{tab:semantic_sft}).

    \item \textbf{Physics–Voxel Policy Optimization.}
    We further introduce PVPO, a physics-informed reinforcement learning framework that couples simulation-based physical-validity rewards with structure-aware geometric rewards.
    Under test-time scaling, PVPO consistently outperforms the model trained on full-scale dataset on structural and semantic alignment (See \Cref{fig:passk-vision}) and whole structure stability (See \Cref{fig:pvpo-compute-ablation}).
    \Cref{fig:confidence} shows that PVPO improves confidence calibration, making high-confidence test-time selections more predictive of true semantic and structural quality.
    PVPO-trained DeepSeek-Llama-8B outperforms DeepSeek-V4, and Kimi-K3 on LBA (See \Cref{fig:case10-frontier}). 

\end{itemize}
\endgroup

\begin{table*}[t]
    \centering
    \scriptsize
    \newcommand{\heat}[2]{\cellcolor{blue!#1}{#2}}
    \newcommand{\mheat}[2]{\cellcolor{blue!#1}{#2}}
    \setlength{\tabcolsep}{2.4pt}
    \renewcommand{\arraystretch}{1.12}
    \resizebox{\textwidth}{!}{%
    \begin{tabular}{lcccccccccccccccc}
      \textbf{Setting} & \multicolumn{8}{c}{\textbf{Qwen2.5-3B-Instruct}} & \multicolumn{8}{c}{\textbf{Llama-3.2-1B-Instruct}}\\
      \shline
      & \textbf{MiMo $\uparrow$} & \textbf{Gemma $\uparrow$} & \textbf{Qwen-VL $\uparrow$} & \textbf{CLIP $\uparrow$} & \textbf{DINOv3 $\uparrow$} & \textbf{Physics $\uparrow$} & \textbf{Voxel $\uparrow$} & \textbf{Bricks}
      & \textbf{MiMo $\uparrow$} & \textbf{Gemma $\uparrow$} & \textbf{Qwen-VL $\uparrow$} & \textbf{CLIP $\uparrow$} & \textbf{DINOv3 $\uparrow$} & \textbf{Physics $\uparrow$} & \textbf{Voxel $\uparrow$} & \textbf{Bricks} \\
      Full dataset & \mheat{48}{0.68} & \mheat{35}{0.39} & \mheat{33}{0.59} & \mheat{43}{0.26} & \mheat{33}{0.67} & \heat{50}{0.93} & \heat{48}{0.32} & 196 & \mheat{54}{0.70} & \mheat{51}{0.56} & \mheat{52}{0.67} & \mheat{52}{\textbf{0.27}} & \mheat{50}{0.74} & \heat{57}{0.96} & \heat{33}{0.35} & 177 \\
      Diversity-only & \mheat{36}{0.52} & \mheat{34}{0.38} & \mheat{31}{0.58} & \mheat{43}{0.26} & \mheat{31}{0.66} & \heat{60}{\textbf{0.95}} & \heat{48}{0.32} & 163 & \mheat{47}{0.62} & \mheat{42}{0.47} & \mheat{39}{0.55} & \mheat{35}{0.25} & \mheat{31}{0.66} & \heat{51}{0.94} & \heat{26}{0.31} & 199 \\
      Random subset & \mheat{34}{0.50} & \mheat{27}{0.30} & \mheat{28}{0.56} & \mheat{35}{0.25} & \mheat{27}{0.64} & \heat{40}{0.91} & \heat{33}{0.28} & 176 & \mheat{41}{0.54} & \mheat{35}{0.39} & \mheat{43}{0.58} & \mheat{35}{0.25} & \mheat{34}{0.67} & \heat{54}{0.95} & \heat{28}{0.32} & 194 \\
      Low-value VLM & \mheat{23}{0.35} & \mheat{19}{0.20} & \mheat{21}{0.51} & \mheat{10}{0.22} & \mheat{27}{0.64} & \heat{10}{0.85} & \heat{22}{0.25} & 144 & \mheat{34}{0.45} & \mheat{28}{0.32} & \mheat{10}{0.28} & \mheat{10}{0.22} & \mheat{10}{0.57} & \heat{31}{0.87} & \heat{23}{0.29} & 334 \\
      Shortest responses & \mheat{10}{0.18} & \mheat{10}{0.10} & \mheat{19}{0.50} & \mheat{27}{0.24} & \mheat{29}{0.65} & \heat{40}{0.91} & \heat{10}{0.22} & 33 & \mheat{10}{0.15} & \mheat{10}{0.13} & \mheat{33}{0.49} & \mheat{27}{0.24} & \mheat{22}{0.62} & \heat{36}{0.89} & \heat{10}{0.22} & 38 \\
      Lowest perplexity & \mheat{18}{0.28} & \mheat{14}{0.15} & \mheat{12}{0.45} & \mheat{35}{0.25} & \mheat{10}{0.56} & \heat{25}{0.88} & \heat{10}{0.22} & 136 & \mheat{45}{0.59} & \mheat{46}{0.51} & \mheat{49}{0.64} & \mheat{52}{\textbf{0.27}} & \mheat{50}{0.74} & \heat{60}{\textbf{0.97}} & \heat{24}{0.30} & 140 \\
      Longest responses & \mheat{29}{0.43} & \mheat{26}{0.29} & \mheat{10}{0.44} & \mheat{18}{0.23} & \mheat{12}{0.57} & \heat{15}{0.86} & \heat{25}{0.26} & \textbf{346} & \mheat{36}{0.48} & \mheat{29}{0.33} & \mheat{32}{0.48} & \mheat{35}{0.25} & \mheat{22}{0.62} & \heat{42}{0.91} & \heat{17}{0.26} & \textbf{351} \\
      \rowcolor{Gray} \textbf{High-Value VLM} & \mheat{55}{0.78} & \mheat{58}{0.66} & \mheat{49}{0.70} & \mheat{52}{0.27} & \mheat{43}{0.72} & \heat{15}{0.86} & \heat{41}{0.30} & 162 & \mheat{53}{0.69} & \mheat{51}{0.56} & \mheat{56}{0.70} & \mheat{52}{\textbf{0.27}} & \mheat{46}{0.72} & \heat{10}{0.80} & \heat{17}{0.26} & 205 \\
      \rowcolor{Gray} \textbf{High-Value VLM + Diversity} & \mheat{60}{0.84} & \mheat{56}{0.63} & \mheat{52}{0.72} & \mheat{43}{0.26} & \mheat{39}{0.70} & \heat{15}{0.86} & \heat{45}{0.31} & 184 & \mheat{53}{0.69} & \mheat{52}{0.57} & \mheat{60}{\textbf{0.74}} & \mheat{52}{\textbf{0.27}} & \mheat{55}{\textbf{0.76}} & \heat{36}{0.89} & \heat{28}{0.32} & 181 \\
      \rowcolor{Gray} \textbf{PVPO} & \mheat{60}{\textbf{0.84}} & \mheat{60}{\textbf{0.68}} & \mheat{60}{\textbf{0.77}} & \mheat{60}{\textbf{0.28}} & \mheat{60}{\textbf{0.80}} & \heat{50}{0.93} & \heat{60}{\textbf{0.35}} & 146 & \mheat{60}{\textbf{0.78}} & \mheat{60}{\textbf{0.66}} & \mheat{60}{\textbf{0.74}} & \mheat{60}{\textbf{0.28}} & \mheat{60}{\textbf{0.78}} & \heat{51}{0.94} & \heat{60}{\textbf{0.50}} & 154 \\
    \end{tabular}%
    }
    \caption{\textbf{Comparisons of Data Selection}: Structure or semantic alignment (Qwen-VL/CLIP/DINOv3), physics validity, voxel alignment, and generated-brick statistics for models trained with different data-selection strategies.}
    \label{tab:semantic_sft}
\end{table*}

\begin{table}[t]
    \centering
    \newcommand{\dheat}[2]{\cellcolor{blue!#1}{#2}}
    \setlength{\tabcolsep}{2.5pt}
    \renewcommand{\arraystretch}{1.12}
    \resizebox{\columnwidth}{!}{%
    \begin{tabular}{lccccccc}
      \textbf{Setting} & \multicolumn{7}{c}{\textbf{DeepSeek-R1-Distill-Llama-8B}} \\
      \shline
      & \textbf{MiMo $\uparrow$} & \textbf{Gemma $\uparrow$} & \textbf{Qwen-VL $\uparrow$} & \textbf{DINOv3 $\uparrow$} & \textbf{CLIP $\uparrow$} & \textbf{Physics $\uparrow$} & \textbf{Voxel $\uparrow$} \\
      Full dataset & \dheat{10}{0.61} & \dheat{10}{0.56} & \dheat{10}{0.67} & \dheat{10}{0.67} & \dheat{10}{0.27} & \dheat{52}{0.991} & \dheat{27}{0.33} \\
      \rowcolor{Gray} \textbf{High-Value VLM + Diversity} & \dheat{16}{0.63} & \dheat{39}{0.60} & \dheat{60}{0.73} & \dheat{60}{0.79} & \dheat{10}{0.27} & \dheat{10}{0.953} & \dheat{60}{\textbf{0.35}} \\
      \rowcolor{Gray} \textbf{PVPO} & \dheat{60}{\textbf{0.80}} & \dheat{60}{\textbf{0.63}} & \dheat{52}{\textbf{0.72}} & \dheat{60}{\textbf{0.79}} & \dheat{60}{\textbf{0.28}} & \dheat{60}{\textbf{0.998}} & \dheat{10}{0.32} \\
    \end{tabular}%
    }
    \caption{Performance of DeepSeek-R1-Distill-Llama-8B on semantic alignment, physics validity, and voxel alignment.}
    \label{tab:deepseek}
\end{table}
\vspace{-3mm}
\section{\physhack{}: Physics--Semantics Misalignment in LBA}
\label{sec:physhack}
We identify \physhack{}, a misalignment phenomenon in LLM-based LBA, where models achieve high physical validity while failing to preserve the intended object semantics. 
\subsection{Preliminaries}
\paragraph{Language Modeling for Brick Assembly.}
Following \citet{pun2025generating,kulits2026bricknet}, we represent each LEGO construction as an executable assembly program generated autoregressively by LLM.
Given a text prompt $x$, the LLM policy $\pi_\theta$ produces a sequence of brick commands $o=(b_1,\ldots,b_T)$:
\begin{equation}
\small
\pi_\theta(o \mid x)
=
\prod_{t=1}^{T}
\pi_\theta\left(b_t \mid x, b_{<t}\right).
\end{equation}

\noindent Each command $b_t$ specifies the brick type and 3D voxel placement of one LEGO brick, and is serialized into 10 tokens.
The generated program is then rendered into 3D brick structure via StableLego simulator \citep{liu2024stablelego}.

\paragraph{Physical Validity.}
Physical validity measures whether a generated LEGO program can be instantiated as a feasible brick assembly. 
For bricks $o=(b_1,\ldots,b_T)$, each $b_i$ must satisfy low-level constraints, including valid brick type, bounded placement, collision-free occupancy, etc.
Let $P(o)\in[0,1]$ denote the physical-validity score of $o$, where a higher value indicates that more generated bricks satisfy these constraints.
However, physical validity alone does not guarantee semantic or structural correctness. 
\vspace{-2mm}
\subsection{Measuring Physics-Semantics Misalignment}
\begin{figure}[!t]
  \centering
  \includegraphics[width=0.90\linewidth]{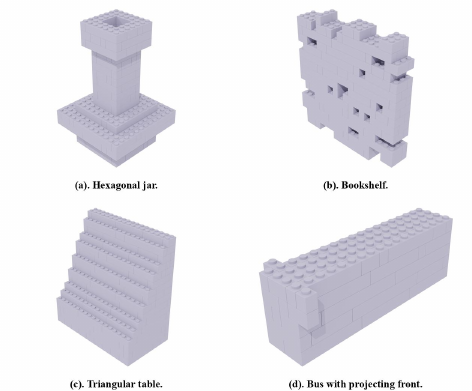}
  \vspace{2pt}
  \resizebox{\linewidth}{!}{%
  \begin{tabular}{lcccccc}
    \textbf{Score Interval} & \textbf{[0,0.5]} & \textbf{(0.5,0.8]} & \textbf{(0.8,0.85]} & \textbf{(0.85,0.9]} & \textbf{(0.9,0.95]} & \textbf{(0.95,1]}\\
    \shline
    \textbf{Percentage (\%)} & 3.52 & 25.06 & 64.63 & 6.80 & 0.00 & 0.00 \\
  \end{tabular}%
  }
  \caption{Physically valid LEGO that not match the intended semantics (The four examples with Qwen-VL score less than 0.4).}
  \label{fig:origin_image}
\end{figure}

This gap is already evident under full-data training (\Cref{tab:semantic_sft}). Although all training trajectories pass physical-validity checks, the resulting Qwen model achieves $0.93$ Validity@4 but performs substantially worse in semantic and structural alignment, reaching only $0.39$ Gemma@4 and $0.67$ DINOv3@4. \Cref{fig:origin_image} reveals a dataset-level source of this behavior. Even among physically feasible trajectories, semantically faithful examples are sparse, and many assemblies omit object-defining components, such as jars without container bodies, etc. Reinforcement learning further amplifies this shortcut: with physical validity as the sole reward to optimize ($\lambda=0$), the model reaches Validity@4 to $1.00$, while semantic and geometric performance remains substantially lower (\Cref{fig:pvpo-compute-ablation}). Together, these results suggest that \physhack{} is induced by physically valid but semantically noisy supervision.

\section{Value-Guided Data Selection for Sample-Efficient Post-Training}
\label{sec:data_select}
To investigate data patterns mitigate \physhack{} and enable efficient yet effective post-training, we treat data selection as a controlled testbed of how different supervision signals shape the learned policy.

\subsection{Experimental Settings}
\paragraph{Backbones and Baselines.}
We adopt following backbones: Qwen2.5-3B-Instruct~\citep{qwen2024qwen2}, Llama-3.2-1B-Instruct~\citep{grattafiori2024llama}, and DeepSeek-R1-Distill-Llama-8B (DPSK-Llama)~\citep{deepseekai2025deepseekr1incentivizingreasoningcapability}.
We compare against five baselines: \textbf{Full dataset}, using all raw data from \citet{pun2025generating}, \textbf{Perplexity}, selecting responses with the lowest perplexity under the full-data model, \textbf{Length}, selecting the longest or shortest responses, \textbf{Diversity-only}, sampling uniformly across domains, and \textbf{Low-value VLM}, selecting examples with the lowest VLM alignment scores, serves as a lower-bound ablation.
The subset-based methods use only $5\%$ of the original training pool, while all other training hyperparameters are held constant across methods.
We train all models with LoRA~\citep{hu2022lora} using rank $32$, a cutoff length of $4{,}096$, and $6$ epochs on $8\times$ NVIDIA RTX 4090 GPUs.

\paragraph{Evaluation Metrics.}
For each prompt and random seed, we sample $K=4$ programs and report Mean@4 metrics averaged across samples, prompts, and four random seeds. 
For semantic and visual-structural alignment, \textbf{MiMo-VL}~\citep{coreteam2025mimovltechnicalreport}, \textbf{Gemma}~\citep{gemmateam2026gemma4}, and \textbf{Qwen-VL}~\citep{bai2023qwen} scores prompt-structure consistency in object identity, attributes, and spatial layout, \textbf{CLIP}~\citep{radford2021learning} measures global image--text alignment, and \textbf{DINOv3}~\citep{simeoni2025dinov3} measures visual structural similarity to the GT reference. 
\textbf{Bricks@4} records the average number of generated bricks. 
The evaluation set is identical from \citet{pun2025generating}.

\subsection{Value-Guided Trajectory Selection} 
For each trajectory $\tau_i=(x_i,o_i)$, where $x_i$ is the text description and $o_i$ is the executable LEGO program, we render $o_i$ into an image via Blender (Appendix \Cref{sec:blender}) and use Qwen2.5-VL as a value model to score text--structure consistency:
\begin{equation}
\small
V(\tau_i)
=
S_{\mathrm{sem}}\bigl(x_i,\mathrm{Render}(o_i)\bigr).
\end{equation}
We formulate trajectory selection as an optimization problem to select a compact and diverse high-value subset:
\begin{equation}
\small
\begin{aligned}
\mathcal{S}^{\star}
=&\;\arg\max_{\mathcal{S}\subseteq\mathcal{D}}
\sum_{\tau_i\in\mathcal{S}} V(\tau_i)
+
\lambda \mathrm{Div}(\mathcal{S}) \\
\mathrm{s.t.}\quad
& P(o_i)\ge\epsilon_{\mathrm{phys}},\ \forall \tau_i\in\mathcal{S},
\quad
|\mathcal{S}|=\rho|\mathcal{D}|.
\end{aligned}
\end{equation}
Here, $P(o_i)$ denotes physical validity, $\rho=0.05$ is the selection ratio, and $\mathrm{Div}(\mathcal{S})$ promotes domain coverage to avoid over-represented categories. 
In practice, we use domain-stratified top-$K$ selection: within each domain, we rank physically valid trajectories by $V(\tau_i)$ and select the top examples under a fixed budget. 
This produces a $20\times$ smaller training set that preserves text--structure consistency, physical feasibility, and domain diversity.

\subsection{Evaluation and Analysis}

\paragraph{Small high-value subsets can outperform full-scale training.}
Tables~\ref{tab:semantic_sft} and~\ref{tab:deepseek} show that compact high-value subsets achieve stronger semantic and structural alignment than full-data training. 
Using only $5\%$ data, \textbf{High-Value VLM + Diversity} improves MiMo@4 from $0.68$ to $0.84$, Gemma@4 from $0.39$ to $0.63$, and Qwen-VL@4 from $0.59$ to $0.72$ on Qwen,
It improves Gemma@4 from $0.56$ to $0.60$, MiMo@4 from $0.61$ to $0.63$ for DPSK-Llama.
It improves DINOv3@4 from $0.67$ to $0.70$ for Qwen, $0.74$ to $0.76$ for Llama, and $0.67$ to $0.79$ for DPSK-Llama, respectively. 
These results suggest that LBA post-training depends more on trajectory value than raw data scale.

\paragraph{Different selection signals reveal complementary data properties.}
The alternative selectors reveal complementary signals for LBA post-training. 
\textbf{Diversity-only} yields the strongest Qwen physical validity ($0.95$ Validity@4), but its Qwen-VL@4 and DINOv3@4 remain below High-Value VLM + Diversity, showing that coverage alone is insufficient. \textbf{Perplexity-based} selection captures syntactically regular programs and performs well on Llama, but transfers poorly to Qwen ($0.45$ Qwen-VL@4). \textbf{Length-based} selection shows that assembly complexity matters: shortest and longest responses produce very different brick counts ($33$ vs. $346$ on Qwen), yet neither extreme yields strong semantic alignment. 
\textbf{Low-value VLM} serves as a negative ablation, dropping Llama Qwen-VL@4 to $0.28$. 
Overall, useful LBA data requires both structural coverage and semantic alignment.

\begin{figure*}[t]
    \centering
    \includegraphics[width=0.95\textwidth]{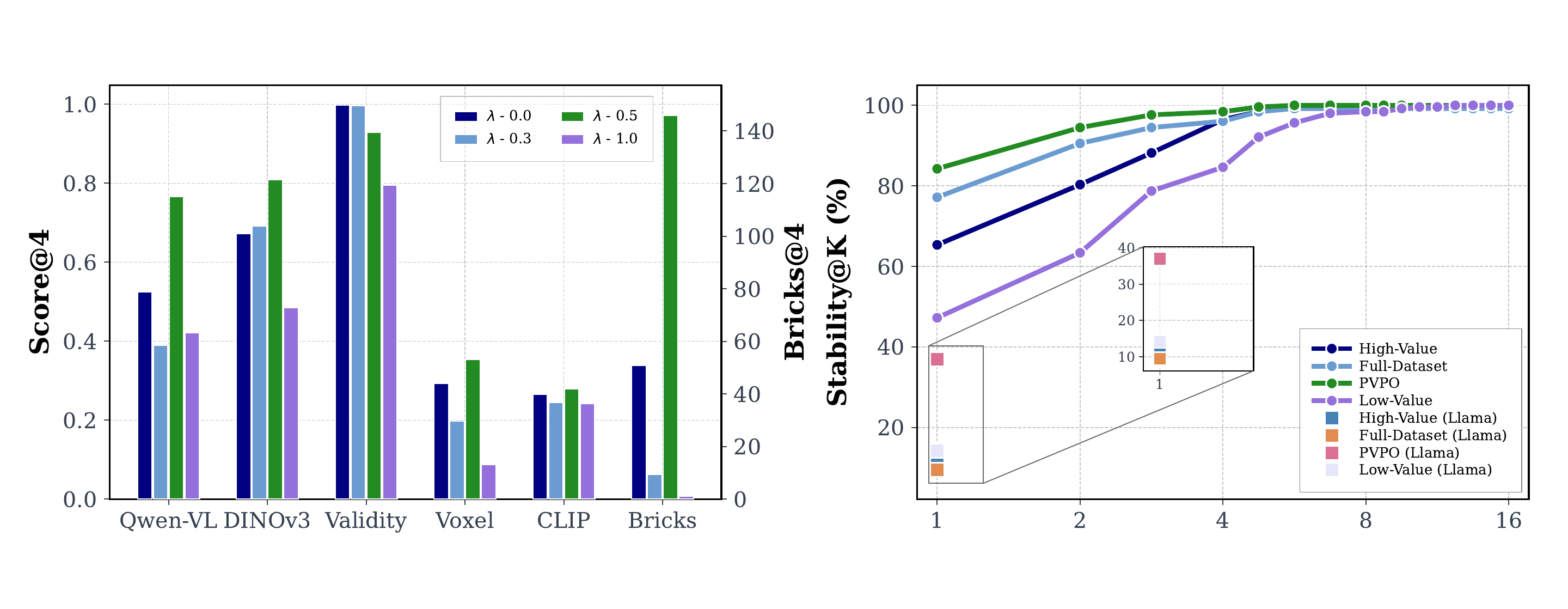}
    \vspace{-1.5em}
    \caption{ \textbf{Left}: Semantic alignment (Qwen-VL/CLIP/DINOv3), physics validity, voxel alignment, and generated-brick number under different voxel weight $\lambda$ on Qwen2.5-3B-Instruct. \textbf{Right}: Stability@K (\%) versus regeneration attempts $K$ during the test-time inference on Qwen2.5-3B-Instruct (K=1--16) and Llama3.2-1B-Instruct (K=1).}
    \label{fig:pvpo-compute-ablation}
\end{figure*}

\begin{figure*}[t]
    \centering
    \includegraphics[width=0.95\textwidth]{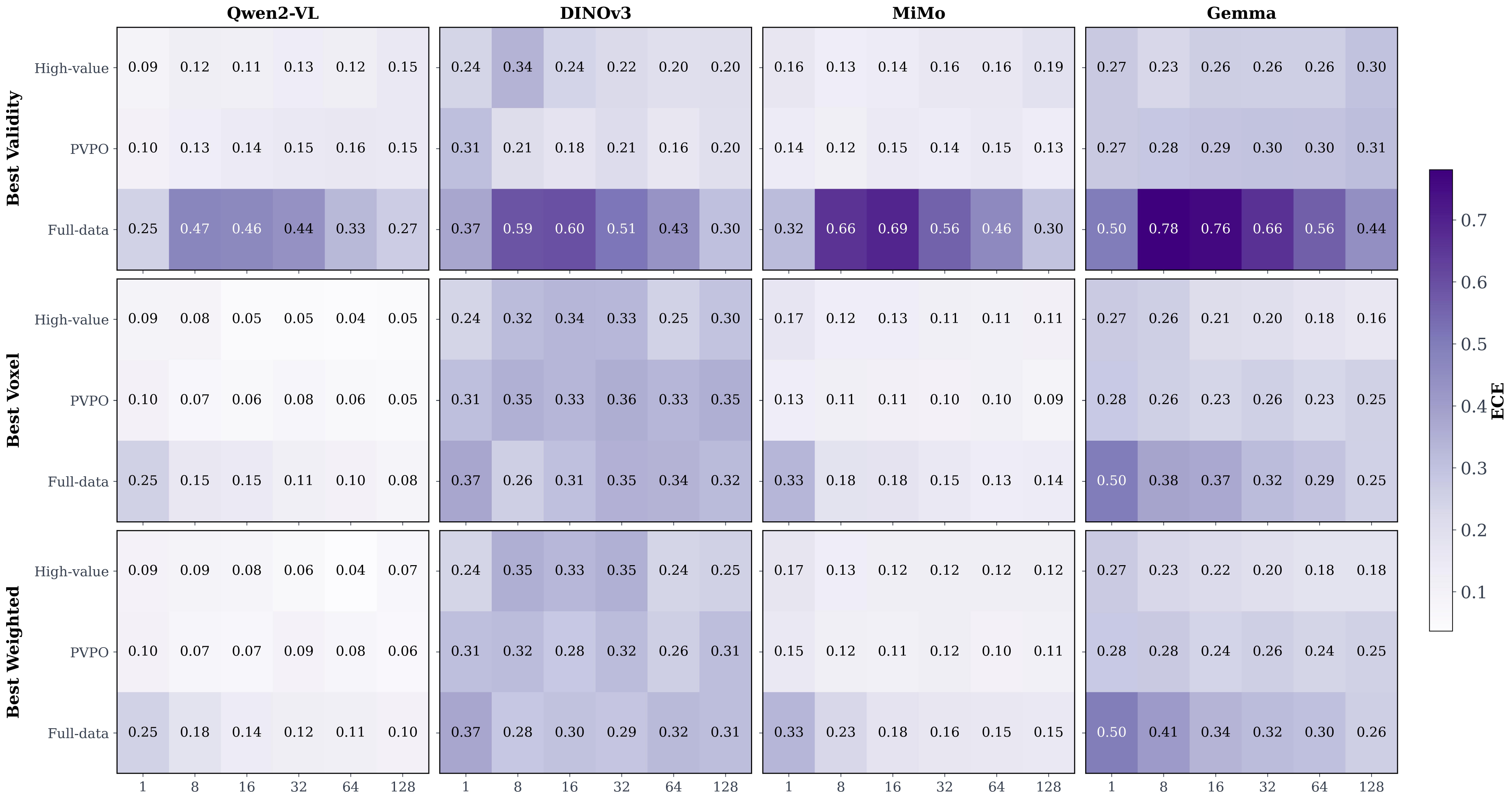}
    \caption{Confidence calibration measured by ECE under three best@k selection mechanisms. Columns blocks correspond to confidence calibrated against Qwen2-VL, DINOv3, MiMo, and Gemma semantic alignment scores and rows correspond to Best Validity, Best Voxel, and Best Weighted selection. Within each block, rows compare High-value, PVPO, and Full-data methods across different inference test-time $K$ (Qwen2.5-3B-Instruct).}
    \label{fig:confidence}
\end{figure*}
\section{PVPO: Exploiting Physics--Structure Consistency via Reinforcement Learning}
\label{sec:rl}
We introduce \emph{Physics--Voxel Policy Optimization} (PVPO), a physics- and structure-aware RL framework for LLM-based LBA. 
Although data selection removes explicitly misaligned trajectories, models may still exploit implicit dataset shortcuts, such as generic brick patterns, length distributions, base structures, or brick-type biases, that satisfy physical validity without preserving object-level semantic correctness~\citep{macdiarmid2025natural,cloud2025subliminal,bowman2022measuring}.

\subsection{Reward Modeling}
\label{sec:reward_modeling}
\paragraph{Modeling Physical Validity as Reward.}
Given a generated program $o=(b_1,\ldots,b_T)$, let $v_t\in\{0,1\}$ indicate whether brick $b_t$ satisfies the required assembly constraints. 
We define physical validity as reward:
\begin{equation}
\small
    R_{\mathrm{phys}}(o)=\frac{1}{T}\sum_{t=1}^{T}v_t,
\end{equation}
to measure the fraction of valid bricks.

\paragraph{Voxel-Space Geometric Alignment Reward.}
We measure geometric alignment in voxel space using Chamfer distance.
Let $\mathcal{V}(o)$ and $\mathcal{V}(o^\star)$ denote the occupied voxel sets of the generated program $o$ and the target construction $o^\star$, respectively. 
We compute a symmetric voxel-space Chamfer distance:
\begin{equation}
\small
\begin{aligned}
D_{\mathrm{vox}}(o,o^\star)
=
\frac{1}{2}\Bigg(
&\frac{1}{|\mathcal{V}(o)|}
\sum_{u\in\mathcal{V}(o)}
\min_{v\in\mathcal{V}(o^\star)}
\|u-v\|_2^2 \\
&\hspace{-2.2em}
+\frac{1}{|\mathcal{V}(o^\star)|}
\sum_{v\in\mathcal{V}(o^\star)}
\min_{u\in\mathcal{V}(o)}
\|u-v\|_2^2
\Bigg)
\end{aligned}
\end{equation}
We normalize this distance and convert it into a reward:
\begin{equation}
\small
R_{\mathrm{vox}}(o,o^\star)
=
1-
\min\left(
\frac{D_{\mathrm{vox}}(o,o^\star)}{d_{\max}},
1
\right)
\end{equation}
where $d_{\max}$ is a normalization constant. 
Although this reward does not directly measure text-level semantics, the target construction $o^\star$ is conditioned on the input prompt, thus voxel reward provides a geometry-level signal for recovering the intended object shape and layout. 

\paragraph{Coupled PVPO Reward.}
PVPO combines physical feasibility and voxel-space structural alignment:
\begin{equation}
\small
R_{\mathrm{PVPO}}(o,o^\star)
=
(1-\lambda)R_{\mathrm{phys}}(o)
+
\lambda R_{\mathrm{vox}}(o,o^\star)
\end{equation}
where $\lambda_{\mathrm{vox}}$ controls the strength of the structure-aware reward. 
In our main experiments, we set $\lambda_{\mathrm{vox}} = 0.5$, which empirically yields the best performance (see \Cref{fig:pvpo-compute-ablation}).

\subsection{RL Training Settings}
We optimize PVPO with a GRPO-style training setup, following \citep{schulman2017proximal,yu2026dapo,shao2024deepseekmath,liu2025understanding,cui2025entropy}.
KL coefficient set to $0.001$ to regularize the online policy against the frozen SFT reference policy. 
Entropy regularization is set to $0.0$, we empirically find that larger entropy weights, such as $0.1$ or $0.2$, lead to policy collapse.
We adopt token-level policy-gradient aggregation. 
To improve rollout utilization, each update batch mainly uses samples from the current policy, with a small portion reused from the previous policy via a replay buffer.
For efficient training, we use LoRA with rank 32 and update only the adapter parameters. The RL stage uses the same dataset as SFT.

\begin{figure*}[t]
    \centering
    \includegraphics[width=0.95\textwidth]{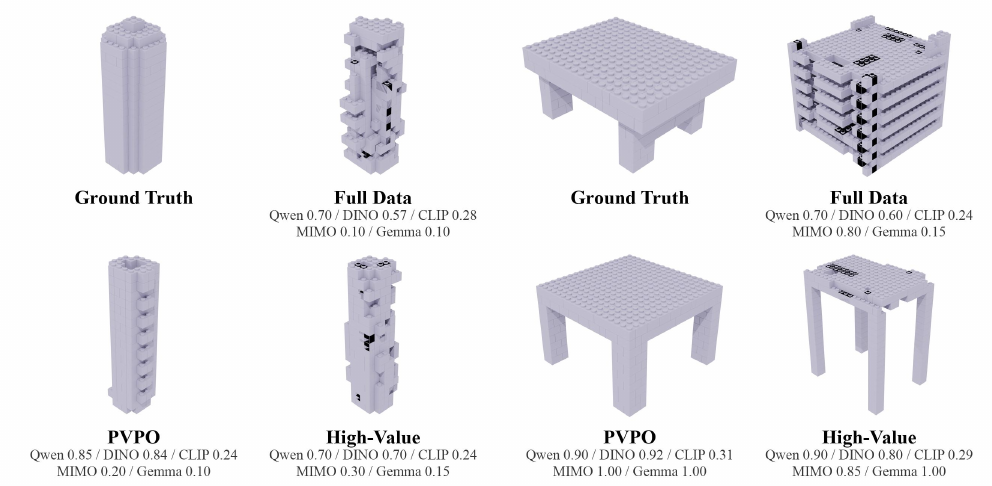}
    \caption{\textbf{Qualitative comparison on two representative LEGO generation tasks}. Bottle (left) and square table (right) show the generated structures from Full Data, PVPO, High-Value training, and ground truth. PVPO and High-Value produce cleaner geometry and better visual alignment than Full Data, black bricks indicate collisions.}
    \label{fig:example_image}
\end{figure*}

\subsection{Evaluation and Analysis}
\paragraph{PVPO improve physics--structure consistency.}
Table~\ref{tab:semantic_sft} and \Cref{tab:deepseek} show that PVPO improves the balance among semantic alignment, geometric fidelity, and physical validity across model backbones. 
On Qwen, compared with full-data training, PVPO increases MiMo@4 from $0.68$ to $0.84$, Gemma@4 from $0.39$ to $0.68$, and DINOv3@4 from $0.67$ to $0.80$. 
It maintains strong physical validity ($0.93$ Validity@4), improves Voxel@4 from $0.32$ to $0.35$, and reduces the average number of generated bricks from $196$ to $146$. 
It suggests that PVPO does not merely increase construction complexity, rather, the coupled reward promotes more compact and structurally faithful assemblies.

\paragraph{PVPO improves semantic alignment under Best@K test-time scaling.}
\Cref{fig:passk-vision} shows that PVPO's alignment advantage persists under Best@K test-time scaling.
For a controlled comparison, each selection criterion is applied identically to all methods, and the selected candidates are evaluated via independent semantic and visual-structural evaluators. 
Under best-by-validity and best-by-voxel selection, the full-data model remains substantially weaker than PVPO in MiMO, Gemma, and DINOv3 scores across most values of $K$, indicating that high physical validity alone does not reliably identify semantically faithful candidates. 
As voxel is used only as a shared candidate-selection criterion, whereas the selected outputs are assessed by independent downstream evaluators, these results suggest that PVPO improves proxy--quality alignment, making test-time selection more predictive of semantic and structural quality.
\Cref{fig:example_image} provides qualitative examples illustrating this trend.

\paragraph{Balanced reward coupling outperforms single-proxy optimization.}
Optimizing either reward alone leads to clear failure modes. 
With $\lambda=0$, physics-only training reaches near-perfect Validity@4 ($1.00$), but yields much lower Qwen-VL@4, DINOv3@4, and Voxel@4 scores ($0.52$, $0.67$, and $0.29$), suggesting that physical feasibility alone can amplify the \physhack{}.
With $\lambda=1.0$, voxel-only training also underperforms: Validity@4 drops to $0.79$, while Qwen-VL@4, DINOv3@4, and Voxel@4 reach only $0.42$, $0.49$, and $0.08$, respectively. 
This suggests that voxel reward becomes unreliable without physical-validity constraints.
In contrast, the balanced setting $\lambda=0.5$ achieves the best overall trade-off, with the strongest Qwen-VL@4 ($0.77$), DINOv3@4 ($0.80$), Voxel@4 ($0.35$), and CLIP@4 ($0.28$), while preserving high Validity@4 ($0.93$). 
These results show that PVPO benefits from coupling physical feasibility with geometric alignment, rather than maximizing either component in isolation.

\paragraph{PVPO improves stability under test-time scaling.}
Repeated inference can improve output quality but increases inference-time compute~\citep{levi2025simple}. \Cref{fig:pvpo-compute-ablation} shows that PVPO achieves high whole-structure stability with few rejection samples (details in \Cref{sec:physics_stable}), increasing from roughly $85\%$ Stability@1 to $95\%$ Stability@2 and approaching saturation by $K=4$. In contrast, the low-value model starts at about $40\%$ Stability@1 and requires substantially more samples to reach comparable performance. Thus, PVPO induces structural stability even without directly optimizing it.

\section{Intriguing Insights and Discussion}
\paragraph{PVPO-Trained DeepSeek-Llama-8B Outperforms Frontier Models.}
\Cref{fig:frontier-grouped-error-bars} shows that PVPO-trained DeepSeek-Llama-8B outperforms frontier models, including Kimi-K3 \citep{kimiteam2026kimik3openfrontier} and the DeepSeek-V4 variants \citep{wang2022translating}, in semantic alignment (MiMo and Qwen-VL), structural similarity (DINOv3), and physical validity. Notably, this advantage persists even when Kimi-K3 is provided with a ground-truth image reference under few-shot strategies.

\paragraph{PVPO Calibrates LLMs for Reliable Physical Reasoning.}
\Cref{fig:confidence} evaluates whether the LLM's self-reported confidence is calibrated with downstream semantic and structural quality. For each selected generation $i$, let $c_i\in[0,1]$ denote model confidence and $s_i\in[0,1]$ the normalized score from an independent evaluator. We compute ECE~\cite{guo2017calibration} across Qwen2-VL, MiMo, Gemma, and DINOv3 under best-by-validity, best-by-voxel, and best-weighted selection, lower ECE indicates better calibration.
Full-data training is strongly miscalibrated under best-by-validity selection: at $K=8$--$16$, its ECE reaches $0.46$--$0.47$ for Qwen2-VL and $0.59$--$0.60$ for DINOv3, compared with $0.13$--$0.14$ and $0.18$--$0.21$ for PVPO. Voxel-based selection improves Qwen2-VL calibration but remains less reliable for DINOv3, whereas best-weighted selection provides a more balanced result, with PVPO's Qwen2-VL ECE decreasing from $0.10$ to $0.06$. Overall, PVPO mitigates \physhack{} by aligning confidence and selection proxies more closely with semantic and structural quality.

\section{Related Works and Concluding Remarks}

\paragraph{LLM for Symbolic, Vision, and Physics Reasoning.}
LLMs, agentic workflows \citep{yao2022react,wei2022chain,muennighoff2025s1}, and post-training techniques \citep{shao2024deepseekmath,yu2026dapo,schulman2017proximal,ouyang2022training} have been increasingly adopted for general symbolic, vision, and physics reasoning \citep{johnson2017clevr,zhang2025agentic,alrashedy2025generating,chen2025symbolic,qiu2025can,wang2023voyager,zheng2026voxelcodebench,yu2025generating,rodriguez2026rendering,zhang2025physreason,verma2024evaluating,lilienthal2026reward,yang2025embodiedbench,yang2024physcene,melnik2023benchmarks,bakhtin2019phyre,xue2023phy,cherian2024llmphy,liang2023code}.
LEGO-Brick Assembly (LBA) is one representative setting that jointly tests these capabilities, requiring models to interpret high-level intents, reason over discrete 3D structures, and satisfy physical constraints during generation.

\paragraph{Generative Modeling and Physics Reasoning Pipeline for Brick Assembly.}
LBA poses a challenging task for generative models \citep{vaswani2017attention,ho2020denoising,velivckovic2017graph}, as it requires precise geometric understanding and physics-aware reasoning to synthesize intent-conditioned brick constructions that are both structurally aligned and physically stable \citep{wen2026bricksim,guo2024treesba,ahn2022budget,ge2024learn,ge2025lego,wang2022translating,thompson2020building,tang2025lego}.
In particular, recent works \citep{pun2025generating,kulits2026bricknet,xu2025legoace,guo2024treesba} adopt pretrained autoregressive language models \citep{radford2019language,grattafiori2024llama,qwen2024qwen2} and formulate LBA as a 3D program synthesis problem under a language modeling framework.
Despite promising progress, several challenges remain in precise physics and vision requirement.
First, physical stability often depends on costly post-hoc rejection sampling~\citep{liu2024statistical}, which does not improve the model's internal understanding of structural feasibility. 
Second, existing methods require substantial human effort for data generation and curation, yet still suffer from frequent structure--text misalignment. 
Moreover, how dataset-level factors affect learning, generalization, and physical reasoning remains poorly understood. 
Finally, reward design for RL-based physical reasoning remains challenging: rewards must support efficient rendering while balancing physical validity, structural feasibility, and downstream quality.

\paragraph{Concluding Remarks}
This work identifies \physhack{} as a key bottleneck in LLM-based LEGO assembly generation. 
We address this challenge with a data-efficient learning framework that combines model-based data selection with PVPO, a sample-efficient RL method coupling physical feasibility and voxel-space structural alignment. 
Our approach calibrates the policy distribution toward generations that are physically valid, stable, compact, and faithful to the prompt.

\section{Limitations}
PVPO has not yet been evaluated in long-horizon agentic LBA settings or integrated into tool harness, where reward shaping and credit assignment may differ substantially from those in the single-turn setting considered in this work, and sandboxed tool use introduces substantial latency during both training and inference.

\bibliography{custom}
\clearpage
\twocolumn
\appendix

\section{Appendix}
\label{sec:appendix}

\subsection{Text-Image Alignment Evaluation}
\paragraph{VLM-Based Text-Image Alignment for Robust Evaluation.} To ensure a controlled comparison and consistent data decontamination, we use the same training and evaluation splits as \citet{pun2025generating}.
We use \textbf{Qwen2.5-VL-7B-Instruct} for data selection and in-distribution evaluation, while \textbf{MiMo-VL-7B-RL} and \textbf{Gemma-4-12B-it} provide independent out-of-distribution evaluations to emperically prove the effectiveness of our data selection and RL methods.
The prompt template used is shown below:

\begin{ptbox}
\footnotesize

\noindent\textbf{Qwen VLM Prompt.} You are an expert evaluator specialized in assessing how well a text description matches a LEGO brick structure image.

Given the image of a LEGO model and the following text description, carefully evaluate the semantic alignment on a scale from 0.0 to 1.0:
\begin{itemize}
    \footnotesize
    \renewcommand\labelitemi{--}
    \setlength{\itemsep}{0pt}
    \item \textbf{1.0}: Perfect match, every important detail in the text is accurately and completely represented in the image.
    \item \textbf{0.9--0.99}: Excellent match, almost perfect, with only very minor visual differences.
    \item \textbf{0.7--0.89}: Good match, main shape, structure, and key features match, but some small details are missing or simplified.
    \item \textbf{0.4--0.69}: Moderate match, core idea is present, but with noticeable discrepancies.
    \item \textbf{0.0--0.39}: Poor match, significant mismatches, missing major elements, or irrelevant content.
\end{itemize}

\noindent\textbf{Text description:} \texttt{\{text\}}

\noindent Output \textbf{ONLY} the following JSON object, nothing else before or after:
\begin{quote}
\footnotesize\ttfamily
\{\{\newline
~~"score": 0.xx,\newline
~~"reason": "One or two clear sentences explaining the main strengths and weaknesses of the match."\newline
\}\}
\end{quote}
\end{ptbox}
\noindent Each model is loaded through the Hugging Face Transformers interface.
Given a rendered LEGO structure and its text description, each VLM is prompted to produce a scalar semantic-alignment score in $[0,1]$. Table\ref{tab:vlm-inference-settings} is the inference settings for the VLM evaluators.
\begin{center}
\begin{minipage}{\columnwidth}
    \centering
    \scriptsize
    \setlength{\tabcolsep}{3pt}
    \renewcommand{\arraystretch}{1.12}
    \resizebox{\columnwidth}{!}{%
    \begin{tabular}{lcccc}
      \textbf{Model} & \textbf{Temperature} & \textbf{Batch Size} & \textbf{Response Length} & \textbf{Dtype} \\
      \shline
      Qwen2.5-VL & 0.0 & 32 & 512 & bf16 \\
      MiMo-VL-7B-RL & 0.0 & 32 & 512 & bf16 \\
      Gemma-4-12B-it & 0.0 & 32 & 512 & bf16 \\
    \end{tabular}%
    }
    \captionof{table}{Inference settings for the VLM evaluators.}
    \label{tab:vlm-inference-settings}
\end{minipage}
\end{center}

\paragraph{CLIP-Based Text-Image Alignment.} We use CLIP \cite{radford2021learning} to measure global text--image alignment. For each rendered LEGO image, the corresponding text prompt is encoded directly without any additional instruction template. The image and text representations are then compared with a CLIP scoring API, which returns a normalized similarity score between the visual representation and the textual description.

\paragraph{Remarks on CLIP Evaluation.}
\label{remark:clip}
The improvement in CLIP alignment achieved by our method is modest compared with its gains on the other evaluation metrics. To contextualize this result, we systematically evaluate CLIP alignment between the ground-truth LEGO structures and their text descriptions in the original training dataset, obtaining an average score of only $0.2954$. This low baseline and narrow score range suggest that CLIP has limited sensitivity in our setting, therefore, we treat small differences in CLIP scores as less informative than improvements in the other evaluation metrics.

\paragraph{Train–evaluation separation.} We follow the official training and evaluation split of BrickGPT \citep{pun2025generating} without modification. 
All data selection, SFT, and RL training are conducted exclusively on the official training split while evaluation is performed on the held-out evaluation split defined by \citep{pun2025generating}.

\paragraph{DINOv3 Image Similarity.} 
For DINOv3-based visual similarity, we use DINOv3 ViT-B/16 \cite{simeoni2025dinov3}.
Rendered LEGO images are encoded with DINOv3, and image-level similarity is computed using cosine similarity between normalized image features. 
When the text branch is enabled, the text description is encoded by the CLIP text encoder, and the DINOv3 image feature is compared with the CLIP text feature after dimensional alignment. Each rendered LEGO image is encoded by a pretrained DINOv3 vision encoder, and the resulting feature vector is L2-normalized. The score is computed as the cosine similarity between the generated image feature and the reference image feature. When multiple reference images are available, we average their DINOv3 features to obtain a single reference representation.

\subsection{Remarks on Automatic Evaluators}
To improve transparency and facilitate qualitative auditing of the automatic
semantic evaluators, we provide results for human evaluation comparing PVPO with Full-data SFTed DeepSeek-Llama-8B.

\paragraph{Human evaluation of semantic alignment.}
We additionally conduct a blinded human evaluation to validate whether the
automatic semantic evaluators agree with human judgments. We randomly sample
50 prompts from the held-out evaluation split and collect one generation from
each of three systems: Full Data SFT, High-Value VLM + Diversity, and PVPO.
This results in 150 text--structure pairs. All structures are rendered using
the same camera, material, lighting, and resolution settings described in
Section~\ref{sec:blender}.

The evaluation is conducted by four annotators who are not involved in the
development of the proposed method. For each example, annotators are shown the
text description and the corresponding rendered LEGO structure. System
identities and automatic evaluation scores are hidden, and the presentation
order is independently randomized for each annotator.

Annotators assess only \textbf{semantic alignment}, since physical validity,
collision avoidance, brick connectivity, and whole-structure stability are
evaluated using deterministic program execution and simulator-based checks.
Semantic alignment measures whether the rendered construction preserves the
object identity, major components, attributes, and spatial organization
specified by the text prompt.

Each example is rated on a five-point Likert scale:

\begin{itemize}
    \item \textbf{1 --- Poor:} The structure does not represent the described
    object, or major object-defining components are missing.
    \item \textbf{2 --- Weak:} The intended object is only vaguely recognizable,
    with substantial semantic or structural mismatches.
    \item \textbf{3 --- Moderate:} The core object is recognizable, but several
    important components, attributes, or spatial relations are inaccurate.
    \item \textbf{4 --- Good:} The main object identity and important components
    are correctly represented, with only minor omissions or simplifications.
    \item \textbf{5 --- Excellent:} The structure faithfully represents the
    object, its major components, attributes, and spatial organization.
\end{itemize}

Before the main annotation, annotators complete a calibration round using
12 examples that are excluded from the final analysis. We average ratings
across annotators and report 95\% confidence intervals obtained with 10,000
bootstrap resamples. 

\paragraph{Human evaluation results.}
Table~\ref{tab:human-semantic-evaluation} reports the human semantic-alignment
ratings. PVPO achieves the highest mean score, followed by High-Value VLM +
Diversity and full-data SFT. This ordering is consistent with the automatic
VLM-based evaluations, indicating that the improvements achieved by PVPO are
also perceptible to human evaluators rather than being specific to a single
automatic scoring model.

\begin{table}[t]
    \centering
    \newcommand{\humanheat}[2]{\cellcolor{blue!#1}{#2}}
    \setlength{\tabcolsep}{2.5pt}
    \renewcommand{\arraystretch}{1.12}
    \resizebox{\columnwidth}{!}{%
    \begin{tabular}{lc}
      \textbf{Setting} & \textbf{Human Semantic Alignment $\uparrow$} \\
      \shline
      Full-data SFT & \humanheat{10}{$2.14 \pm 0.21$} \\
      \rowcolor{Gray} \textbf{PVPO} & \humanheat{60}{$\mathbf{3.06 \pm 0.24}$} \\
    \end{tabular}%
    }
    \caption{Blinded human evaluation of semantic alignment on 50 held-out prompts. Scores are measured on a five-point Likert scale and averaged across four annotators. The 95\% confidence intervals are $[1.94, 2.35]$ for Full-data SFT and $[2.82, 3.30]$ for PVPO.}
    \label{tab:human-semantic-evaluation}
\end{table}
Overall, the human evaluation provides evidence that the automatic VLM
evaluators capture meaningful differences in semantic quality.

\subsection{Rendering Details}
\label{sec:blender}
All LEGO structures are rendered from LDraw (\texttt{.ldr}) files using Blender with the ImportLDraw plugin. We use the Cycles renderer with 64 samples and render each image at a resolution of $2048 \times 2048$. The camera field of view is fixed to $35^\circ$. We use a studio-style setup with a pure white background and no ground plane. All bricks are rendered with a uniform light-purple material, with RGB value $(0.58, 0.48, 0.86)$ and roughness $0.55$. Stud logos are disabled, brick gaps are enabled, and bevels are applied to brick edges with bevel width $0.5$. The LDraw import scale is set to $0.02$. Lighting is provided by two directional sun lights: a key light with energy $2.5$ and a fill light with energy $1.2$. We use Filmic/AgX tone mapping for the bricks and composite the transparent render onto a pure white background. These rendering settings are used only for visualization and do not affect quantitative evaluation.

\paragraph{Why not image-based rewards?}
We also explored using rendered images as references for reward modeling, such as computing visual alignment between generated and target assemblies. 
However, this is computationally prohibitive for reinforcement learning: rendering a single LEGO structure with Blender takes roughly one minute on our hardware, making rollout-time image rendering impractical. 
Therefore, PVPO uses voxel-space geometric alignment as a lightweight structure-aware reward, which can be computed directly from the generated brick occupancy without invoking the rendering pipeline.

\begin{table*}[t]
    \centering
    \scriptsize
    \setlength{\tabcolsep}{2.4pt}
    \renewcommand{\arraystretch}{1.12}
    \resizebox{\textwidth}{!}{%
    \begin{tabular}{lcccccccc}
      \textbf{Setting} & \multicolumn{4}{c}{\textbf{Qwen2.5-3B-Instruct}} & \multicolumn{4}{c}{\textbf{Llama-3.2-1B-Instruct}} \\
      \shline
      & \textbf{\% Valid@200} & \textbf{\% Stable@1} & \textbf{Mean Stability@1} & \textbf{Min Stability}
      & \textbf{\% Valid@200} & \textbf{\% Stable@1} & \textbf{Mean Stability@1} & \textbf{Min Stability} \\
      Full dataset & 99.61 & 77.17 & 0.9986 & 0.9739 & 100.00 & 31.25 & 0.9934 & 0.8890 \\
      Low-value VLM & 100.00 & 47.24 & 0.9983 & 0.9598 & 100.00 & 0.787 & 0.9980 & 0.5952 \\
      \rowcolor{Gray} \textbf{High-Value VLM} & 100.00 & 65.35 & 0.9989 & 0.9736 & 18.90 & 12.60 & 0.9954 & 0.8837 \\
      \rowcolor{Gray} \textbf{PVPO} & 100.00 & \textbf{84.25} & 0.9992 & 0.9802 & 100.00 & \textbf{37.01} & 0.9970 & 0.3329 \\
    \end{tabular}%
    }
    \caption{Structure-level validity rate and stability rate under rejection sampling (N=200) and stability evaluation (K=1).}
    \label{tab:appendix-stability}
\end{table*}
\subsection{Physics-Guided Rejection Sampling and Regeneration}
\label{sec:physics_stable}
To evaluate the Validity of generated LEGO structures, we use a two-stage stability-aware inference procedure following the computation protocol of \citet{pun2025generating}, with summary statistics reported in \cref{tab:appendix-stability}. The first stage performs individual-brick-level rejection sampling, while the second stage applies full-structure-level stability-guided regeneration. In \cref{tab:appendix-stability}, \%Valid@200 denotes the structure-level validity rate with a cumulative budget of 200 rejection-sampling attempts per case. This budget is shared across all brick-generation steps within a case. A structure is considered valid if it is successfully completed with every brick satisfying all validity constraints before the cumulative budget is exhausted. Otherwise, the entire structure is classified as invalid. \%Valid@200 is calculated as the number of valid structures divided by the total number of evaluated cases. \%Stable@1 denotes the percentage of cases that produce a structure that is vaild and stable within a single structure-generation attempt. 

During autoregressive generation, the model predicts one LEGO brick at a time in the format $h \times w\,(x,y,z)$. Each candidate brick is checked for format validity, library membership, grid bounds, collisions, and duplicate invalid proposals. Invalid bricks are rejected, the model state is rolled back, and a new candidate is sampled, with a budget of up to 200 rejections per generated single structure.

After a full structure is generated, we evaluate its physical stability using the BrickGPT stability analyzer. The analyzer produces a voxel-level stability score over the occupied volume of the structure, where larger values indicate greater stability and non-positive values indicate unstable regions. For each brick, we define its brick-level stability as the minimum stability score over all voxels occupied by that brick:
\begin{equation}
\small
s_i = \min_{v \in \mathcal{V}_i} S(v),
\end{equation}
where $S(v)$ denotes the voxel-level stability score and $\mathcal{V}_i$ is the set of voxels occupied by brick $i$. We summarize the structure using the mean and minimum brick stability:
\begin{equation}
\small
S_{\mathrm{mean}} = \frac{1}{N}\sum_{i=1}^{N} s_i,
\qquad
S_{\mathrm{min}} = \min_i s_i,
\end{equation}
where $N$ is the number of generated bricks. A structure is considered physically stable if $S_{\mathrm{min}} > 0$, meaning that even the weakest brick has positive stability.

If the generated structure is unstable, we apply stability-guided regeneration. Specifically, we identify the first unstable brick in the generation order, remove that brick and all subsequent bricks, and keep the remaining stable prefix. The model then continues generation from this stable prefix. This rollback-and-regeneration process can be repeated up to a predefined maximum number of regenerations. In the reported setting, we allow one structure-level regeneration. We also use a tiered regeneration protocol in which samples that already satisfy $S_{\mathrm{min}} > 0$ are frozen and excluded from later regeneration rounds.

This procedure combines local syntactic and geometric filtering with global physical stability checking. Rejection sampling prevents invalid bricks from entering the structure, while regeneration corrects higher-level instability that only becomes apparent after evaluating the assembled model.
\paragraph{Gurobi-based Stability Optimization.}
We compute physical stability using a Gurobi force-equilibrium solver, students can get Gurobi license free. Each generated LEGO structure is converted into a voxelized brick assembly, and contact-force variables are introduced at brick interfaces. The optimizer enforces action-reaction consistency between contacting bricks and minimizes the total residual force and torque imbalance:
\begin{equation}
\small
\resizebox{0.95\columnwidth}{!}{$
\displaystyle
\mathcal{L}_{\mathrm{eq}}
=
\sum_i
\left(
|\Delta F_{x,i}|+
|\Delta F_{y,i}|+
|\Delta F_{z,i}|+
|\Delta \tau_{1,i}|+
|\Delta \tau_{2,i}|
\right).
$}
\end{equation}
We further add small regularization terms on the maximum downward contact force per brick and the total downward contact force:
\begin{equation}
\small
\mathcal{L}
=
\mathcal{L}_{\mathrm{eq}}
+
\alpha \sum_i F^{\max}_{\mathrm{down},i}
+
\beta \sum_j F_{\mathrm{down},j},
\end{equation}
with $\alpha=10^{-3}$ and $\beta=10^{-6}$. The solver uses $g=9.8$, LEGO unit height $0.0096$, unit length $0.0078$, and contact threshold $T=100$, converted to $F_T = T g / 1000$.

After optimization, each occupied voxel is assigned a stability score. If force or torque equilibrium is violated, or if the contact-force margin is non-positive, the voxel score is set to zero. Otherwise, the score is the normalized margin
\begin{equation}
\small
S(v)=\frac{F_T-D_{\max}}{F_T},
\end{equation}
where $D_{\max}$ is the maximum downward contact force. The brick-level stability is the minimum voxel score over the brick, and the structure is considered stable when the minimum brick stability is greater than zero.

\subsection{Dataset}
Training dataset consists of brick-text paired text and brick programs, where each input is a natural-language description of an object and each output is an executable LEGO brick program. 
A brick program is represented as a list of bricks in the format \texttt{<brick dimension> (x,y,z)}, using a fixed library of 14 brick types: \texttt{1x1}, \texttt{1x2}, \texttt{2x1}, \texttt{1x4}, \texttt{4x1}, \texttt{1x6}, \texttt{6x1}, \texttt{1x8}, \texttt{8x1}, \texttt{2x2}, \texttt{2x4}, \texttt{4x2}, \texttt{2x6}, and \texttt{6x2}. 
We consider a full-data SFT setting with 213,020 text-structure pairs from the open-source BrickGPT dataset~\citep{pun2025generating}, which is released under the MIT license. All subset-based SFT settings use 11k examples selected from this full dataset. These subsets are curated with different selection mechanisms, including high-value VLM, high-value VLM + Diversity, low-value VLM, diversity-only, random, longest-response, and shortest-response selection. The RL stage uses a 2K-example subset of the 11K examples selected by the High-value VLM + Diversity strategy. Because the 11K set contains approximately five to six descriptions for each object, we retain only the description with the highest alignment score for each unique object, resulting in approximately 2K object–description pairs for RL training.
For physics-guided reinforcement learning, we use 2k prompts in GRPO format. These prompts are deduplicated and selected as the highest-scoring 2k samples from the 11k high-value subset.
Evaluation is conducted on a deduplicated set of diverse objects designed to emphasize physically meaningful construction principles.

\begin{table}[t]
    \centering
    \scriptsize
    \setlength{\tabcolsep}{3pt}
    \renewcommand{\arraystretch}{1.08}
    \resizebox{\columnwidth}{!}{%
    \begin{tabular}{>{\raggedright\arraybackslash}p{0.30\columnwidth}
                    >{\raggedleft\arraybackslash}p{0.29\columnwidth}
                    >{\raggedleft\arraybackslash}p{0.29\columnwidth}}
      \textbf{Hyperparameter} & \textbf{RL } & \textbf{SFT} \\
      \shline
      Training framework & verl 0.7.0.dev0 & TRL 0.26.0 \\
      Number of epochs & 6 & 6 \\
      Optimizer & AdamW & AdamW \\
      Learning rate & $5\times10^{-6}$ & $2\times10^{-3}$ \\
      Learning-rate warm-up & None & 100 steps \\
      Global batch size & 32 & 64 \\
      PPO mini-batch size & 32 & 64 \\
      Numerical precision & BF16 & BF16 \\
      LoRA($r$) & 32 & 32 \\
      LoRA($\alpha$) & 64 & 64 \\
      Maximum prompt& 300 & / \\
      Maximum response & 3{,}796 & / \\
      Maximum sequence & 4{,}096 & 4{,}096 \\
      Sampling N & 8 & / \\
      Temperature & 1.0 & / \\
      KL-loss coefficient & 0.001 & / \\
      Entropy coefficient & 0 & / \\
      GPUs & $8\times$ NVIDIA RTX 4090 & $8\times$ NVIDIA RTX 4090 \\
    \end{tabular}
    }
    \caption{Training hyperparameters for supervised fine-tuning and PVPO reinforcement learning.}
    \label{tab:training-hyperparameters}
\end{table}

\begin{table}[t]
    \centering
    \scriptsize
    \setlength{\tabcolsep}{3pt}
    \renewcommand{\arraystretch}{1.12}
    \resizebox{\columnwidth}{!}{%
    \begin{tabular}{lll}
      \textbf{Model} & \textbf{Usage} & \textbf{License} \\
      \shline
      Qwen2.5-3B-Instruct & SFT / GRPO & Qwen Research \\
      Llama-3.2-1B-Instruct & SFT / GRPO & Llama 3.2 \\
      DeepSeek-R1-Distill-Llama-8B & SFT / GRPO & MIT \\
      MiMo-VL-7B-RL & VLM evaluation & MIT \\
      Gemma-4-12B-it & VLM evaluation & Apache-2.0 \\
      Qwen2.5-VL-7B-Instruct & VLM evaluation & Apache-2.0 \\
      CLIP ViT-B/32 & VLM evaluation & MIT \\
      DINOv3 ViT-B/16 & VLM evaluation & dinov3-license \\
    \end{tabular}%
    }
    \caption{Model usage and licenses.}
    \label{tab:model-licenses}
\end{table}

\subsection{More About PVPO}
\paragraph{Remarks on Voxel-Space Geometric Alignment Reward of PVPO.}
PVPO uses paired target constructions only to define verifiable rewards during
training, following the reinforcement learning with verifiable rewards
paradigm~\cite{shao2024deepseekmath}. Reference solutions are commonly used in
recent GRPO-style frameworks: rubric-based methods may use ground-truth
solution traces as evidence for process reward models~\citep{tan2026papo},
whereas outcome-reward methods compare generated answers with ground-truth
final answers to assign binary rewards~\citep{shao2024deepseekmath}.
PVPO combines two complementary signals. The physical-validity reward is
verified automatically and does not require access to the target construction,
whereas the voxel-alignment reward is computed against one paired reference
structure. Through on-policy optimization, these signals encourage the model
to learn a general LEGO assembly policy rather than memorize individual target
programs. At inference time, the model receives only the text prompt; neither
the target program nor its voxel representation is available. Moreover,
evaluation is conducted on the clean held-out split of~\citet{pun2025generating},
so the reported gains reflect generalization to unseen prompts rather than
access to reference geometry at train or test time. Because a single text prompt can
admit many semantically valid LEGO constructions, successful generalization
requires the model to learn transferable spatial and physical principles rather
than reproduce a unique reference structure. The consistent improvements on
held-out prompts (on MiMo, Gemma, and human evaluation for all models from Full-data to PVPO) indicate that PVPO acquires a transferable policy,
producing physically feasible and semantically aligned assemblies even when
their exact brick configurations differ from the training references.
PVPO is also fundamentally different from SFT with an additional geometric
loss. The voxel reward is non-differentiable with respect to the discrete LEGO
program and is evaluated only after executing a free-running sampled sequence.
Policy optimization therefore assigns sequence-level credit to complete
assemblies and favors valid, target-aligned trajectories over collapsed or
misaligned ones. Importantly, PVPO requires no geometric annotations beyond
the paired LEGO programs already used for SFT.

\noindent \textbf{Training Hyperparameters.}
The training configurations used for supervised fine-tuning and reinforcement learning are summarized in \cref{tab:training-hyperparameters}.

\begin{table}[t]
    \centering
    \newcommand{\dheatstd}[2]{\cellcolor{blue!#1}{#2}}
    \setlength{\tabcolsep}{2.5pt}
    \renewcommand{\arraystretch}{1.12}
    \resizebox{\columnwidth}{!}{%
    \begin{tabular}{lccc}
      \textbf{Setting} & \multicolumn{3}{c}{\textbf{DeepSeek-R1-Distill-Llama-8B}} \\
      \shline
      & \textbf{MiMo@4 $\uparrow$} & \textbf{Gemma@4 $\uparrow$} & \textbf{Physics@4 $\uparrow$} \\
      Full dataset & \dheatstd{10}{$0.61 \pm 0.22\%$} & \dheatstd{10}{$0.56 \pm 0.11\%$} & \dheatstd{52}{$0.991 \pm 0.049\%$} \\
      \rowcolor{Gray} \textbf{High-Value VLM + Diversity} & \dheatstd{16}{$0.63 \pm 0.92\%$} & \dheatstd{39}{$0.60 \pm 0.63\%$} & \dheatstd{10}{$0.953 \pm 0.175\%$} \\
      \rowcolor{Gray} \textbf{PVPO} & \dheatstd{60}{$\mathbf{0.80 \pm 1.29\%}$} & \dheatstd{60}{$\mathbf{0.63 \pm 1.78\%}$} & \dheatstd{60}{$\mathbf{0.998 \pm 0.019\%}$} \\
    \end{tabular}%
    }
    \caption{Performance of DeepSeek-R1-Distill-Llama-8B, reported as the mean and standard deviation across evaluation runs.}
    \label{tab:deepseek-std}
\end{table}

\subsection{Inference Parameters of Open-Source LLMs: Kimi-K3, DeepSeek-V4 Pro, and DeepSeek-V4 Flash}

\noindent For Kimi-K3 zero-shot evaluation, we retain the dataset's initial system message and the evaluation user message. The complete text-only prompt template is shown below:

\begin{ptbox}
\footnotesize
\noindent\textbf{Kimi 0-Shot without GT Image.}

\medskip
\noindent\textbf{System:}

\noindent You are a helpful assistant.

\medskip
\noindent\textbf{User:}

\noindent Create a LEGO model of the input. Format your response as a list of bricks: \texttt{<brick dimensions> <brick position>}, where the brick position is \texttt{(x,y,z)}.

\noindent Allowed brick dimensions are \texttt{2x4}, \texttt{4x2}, \texttt{2x6}, \texttt{6x2}, \texttt{1x2}, \texttt{2x1}, \texttt{1x4}, \texttt{4x1}, \texttt{1x6}, \texttt{6x1}, \texttt{1x8}, \texttt{8x1}, \texttt{1x1}, and \texttt{2x2}. All bricks are 1 unit tall.

\medskip
\noindent\textbf{Input:}

\begin{quote}
\footnotesize\ttfamily
<text>
\end{quote}

\noindent Use only the text description in the input as the target LEGO model specification. No ground-truth image is provided.

\medskip
\noindent\textbf{Strict output format:}
\begin{itemize}
    \footnotesize
    \renewcommand\labelitemi{--}
    \setlength{\itemsep}{0pt}
    \item Output only brick lines, with no explanation.
    \item Do not write the word ``brick.''
    \item Each line must be exactly: \texttt{<dims> (<x>,<y>,<z>)}.
    \item Example: \texttt{2x4 (1,2,0)}.
    \item The world is $20\times20\times20$, so every coordinate must satisfy $0 \le x,y,z \le 19$.
    \item Never output any brick whose occupied area exceeds the $20\times20\times20$ bounds.
    \item Stop immediately after the last valid brick line.
    \item Do not repeat patterns forever. Do not keep increasing $z$ after the object is complete.
\end{itemize}

\medskip
\hrule
\medskip

\noindent\textbf{Kimi 3-Shot without GT Image.}

\medskip
\noindent\textbf{System:}

\noindent You are a helpful assistant.

\medskip
\noindent\textbf{User:}

\noindent Square basket with layered construction and a recessed interior.

\noindent\textbf{Assistant:}

\noindent "1x1 (15,17,0) 1x4 (15,13,0) 1x8 ...(0,19,5) 1x8 (0,11,5) 1x8 (0,3,5)\textbackslash n"

\noindent\textbf{User:}

\noindent A square-shaped basket featuring stacked horizontal layers and deep inset.

\noindent\textbf{Assistant:}

\noindent "1x1 (15,17,0) 1x4 (15,13,0)1x8 ...(0,11,5) 1x8 (0,3,5)\textbackslash n"

\noindent\textbf{User:}

\noindent Basket with square form, tiered sides, recessed square top, and flat base.

\noindent\textbf{Assistant:}

\noindent "1x1 (15,17,0) 1x4 (15,13,0) 1x8 (15,5,0) ...(0,11,5) 1x8 (0,3,5)\textbackslash n"

\medskip
\noindent\textbf{User:}

\noindent Create a LEGO model of the input. Format your response as a list of bricks: \texttt{<brick dimensions> <brick position>}, where the brick position is \texttt{(x,y,z)}.

\noindent Allowed brick dimensions are \texttt{2x4}, \texttt{4x2}, \texttt{2x6}, \texttt{6x2}, \texttt{1x2}, \texttt{2x1}, \texttt{1x4}, \texttt{4x1}, \texttt{1x6}, \texttt{6x1}, \texttt{1x8}, \texttt{8x1}, \texttt{1x1}, and \texttt{2x2}. All bricks are 1 unit tall.

\medskip
\noindent\textbf{Input:}

\begin{quote}
\footnotesize\ttfamily
<text>
\end{quote}

\noindent Use only the text description in the input as the target LEGO model specification. No ground-truth image is provided.

\medskip
\noindent\textbf{Strict output format:}
\begin{itemize}
    \footnotesize
    \renewcommand\labelitemi{--}
    \setlength{\itemsep}{0pt}
    \item Output only brick lines, with no explanation.
    \item Do not write the word ``brick.''
    \item Each line must be exactly: \texttt{<dims> (<x>,<y>,<z>)}.
    \item Example: \texttt{2x4 (1,2,0)}.
    \item The world is $20\times20\times20$, so every coordinate must satisfy $0 \le x,y,z \le 19$.
    \item Never output any brick whose occupied area exceeds the $20\times20\times20$ bounds.
    \item Stop immediately after the last valid brick line.
    \item Do not repeat patterns forever. Do not keep increasing $z$ after the object is complete.
\end{itemize}

\medskip
\hrule
\medskip

\noindent\textbf{Kimi 0-Shot with GT Image.}

\medskip
\noindent\textbf{System:}

\noindent You are a helpful assistant.

\medskip
\noindent\textbf{User content part 1:}

\begin{quote}
\footnotesize\ttfamily
\{\newline
~~"type": "image\_url",\newline
~~"image\_url": \{"url": "<ground-truth image data url or image url>"\}\newline
\}
\end{quote}

\medskip
\noindent\textbf{User content part 2:}

\noindent Create a LEGO model of the input. Format your response as a list of bricks: \texttt{<brick dimensions> <brick position>}, where the brick position is \texttt{(x,y,z)}.

\noindent Allowed brick dimensions are \texttt{2x4}, \texttt{4x2}, \texttt{2x6}, \texttt{6x2}, \texttt{1x2}, \texttt{2x1}, \texttt{1x4}, \texttt{4x1}, \texttt{1x6}, \texttt{6x1}, \texttt{1x8}, \texttt{8x1}, \texttt{1x1}, and \texttt{2x2}. All bricks are 1 unit tall.

\medskip
\noindent\textbf{Input:}

\begin{quote}
\footnotesize\ttfamily
<eval input text>
\end{quote}

\noindent Use the attached image as the primary geometric reference for the target LEGO model, including its overall shape, proportions, and visible spatial structure.

\medskip
\noindent\textbf{Strict output format:}
\begin{itemize}
    \footnotesize
    \renewcommand\labelitemi{--}
    \setlength{\itemsep}{0pt}
    \item Output only brick lines, with no explanation.
    \item Do not write the word ``brick.''
    \item Each line must be exactly: \texttt{<dims> (<x>,<y>,<z>)}.
    \item Example: \texttt{2x4 (1,2,0)}.
    \item The world is $20\times20\times20$, so every coordinate must satisfy $0 \le x,y,z \le 19$.
    \item Never output any brick whose occupied area exceeds the $20\times20\times20$ bounds.
    \item Stop immediately after the last valid brick line.
    \item Do not repeat patterns forever. Do not keep increasing $z$ after the object is complete.
\end{itemize}

\end{ptbox}

\begin{ptbox}
\footnotesize
\noindent\textbf{DeepSeek-V4 Pro/Flash 0-Shot.}

\medskip
\noindent\textbf{System:}

\noindent You are a helpful assistant.

\medskip
\noindent\textbf{User:}

\noindent Create a LEGO model of the input. Format your response as a list of bricks: \texttt{<brick dimensions> <brick position>}, where the brick position is \texttt{(x,y,z)}.

\noindent Allowed brick dimensions are \texttt{2x4}, \texttt{4x2}, \texttt{2x6}, \texttt{6x2}, \texttt{1x2}, \texttt{2x1}, \texttt{1x4}, \texttt{4x1}, \texttt{1x6}, \texttt{6x1}, \texttt{1x8}, \texttt{8x1}, \texttt{1x1}, and \texttt{2x2}. All bricks are 1 unit tall.

\medskip
\noindent\textbf{Input:}

\begin{quote}
\footnotesize\ttfamily
<eval input text>
\end{quote}

\medskip
\noindent\textbf{Strict output format:}
\begin{itemize}
    \footnotesize
    \renewcommand\labelitemi{--}
    \setlength{\itemsep}{0pt}
    \item Output only brick lines, with no explanation.
    \item Do not write the word ``brick.''
    \item Each line must be exactly: \texttt{<dims> (<x>,<y>,<z>)}.
    \item Example: \texttt{2x4 (1,2,0)}.
    \item The world is $20\times20\times20$, so every coordinate must satisfy $0 \le x,y,z \le 19$.
    \item Never output any brick whose occupied area exceeds the $20\times20\times20$ bounds.
    \item Stop immediately after the last valid brick line.
    \item Do not repeat patterns forever. Do not keep increasing $z$ after the object is complete.
\end{itemize}

\medskip
\hrule
\medskip

\noindent\textbf{DeepSeek-V4 Pro/Flash 3-Shot.}

\medskip
\noindent\textbf{System:}

\noindent You are a helpful assistant.

\medskip
\noindent\textbf{User:}
Square basket with layered construction and a recessed interior.

\noindent\textbf{Assistant:}
"1x1 (15,17,0) 1x4 (15,13,0) 1x8 ...(0,19,5) 1x8 (0,11,5) 1x8 (0,3,5)\textbackslash n"

\noindent\textbf{User:}
A square-shaped basket featuring stacked horizontal layers and deep inset.

\noindent\textbf{Assistant:}
"1x1 (15,17,0) 1x4 (15,13,0)1x8 ...(0,11,5) 1x8 (0,3,5)\textbackslash n"

\noindent\textbf{User:}
Basket with square form, tiered sides, recessed square top, and flat base.

\noindent\textbf{Assistant:}
"1x1 (15,17,0) 1x4 (15,13,0) 1x8 (15,5,0) ...(0,11,5) 1x8 (0,3,5)\textbackslash n"

\noindent\textbf{User:}

\noindent Create a LEGO model of the input. Format your response as a list of bricks: \texttt{<brick dimensions> <brick position>}, where the brick position is \texttt{(x,y,z)}.

\noindent Allowed brick dimensions are \texttt{2x4}, \texttt{4x2}, \texttt{2x6}, \texttt{6x2}, \texttt{1x2}, \texttt{2x1}, \texttt{1x4}, \texttt{4x1}, \texttt{1x6}, \texttt{6x1}, \texttt{1x8}, \texttt{8x1}, \texttt{1x1}, and \texttt{2x2}. All bricks are 1 unit tall.

\medskip
\noindent\textbf{Input:}

\begin{quote}
\footnotesize\ttfamily
<eval input text>
\end{quote}

\medskip
\noindent\textbf{Strict output format:}
\begin{itemize}
    \footnotesize
    \renewcommand\labelitemi{--}
    \setlength{\itemsep}{0pt}
    \item Output only brick lines, with no explanation.
    \item Do not write the word ``brick.''
    \item Each line must be exactly: \texttt{<dims> (<x>,<y>,<z>)}.
    \item Example: \texttt{2x4 (1,2,0)}.
    \item The world is $20\times20\times20$, so every coordinate must satisfy $0 \le x,y,z \le 19$.
    \item Never output any brick whose occupied area exceeds the $20\times20\times20$ bounds.
    \item Stop immediately after the last valid brick line.
    \item Do not repeat patterns forever. Do not keep increasing $z$ after the object is complete.
\end{itemize}
\end{ptbox}

\noindent \textbf{Inference configurations.} The inference configurations are summarized in
Table~\ref{tab:frontier-inference-settings}.
\begin{center}
\begin{minipage}{\columnwidth}
    \centering
    \scriptsize
    \setlength{\tabcolsep}{3pt}
    \renewcommand{\arraystretch}{1.12}
    \begin{tabular}{>{\raggedright\arraybackslash}p{0.28\columnwidth}
                    >{\centering\arraybackslash}p{0.13\columnwidth}
                    >{\centering\arraybackslash}p{0.10\columnwidth}
                    >{\centering\arraybackslash}p{0.15\columnwidth}
                    >{\centering\arraybackslash}p{0.20\columnwidth}}
      \textbf{Model} & \textbf{Temp.} & \textbf{Top-$p$} & \textbf{Max Tok.} & \textbf{Reasoning} \\
      \shline
      Kimi-K3 & 1.0 & 0.95 & 3{,}600 & low \\
      DeepSeek-V4-Flash & 0.6 & 1.0 & 3{,}600 & no \\
      DeepSeek-V4-Pro & 0.6 & 1.0 & 3{,}600 & no \\
    \end{tabular}
    \captionof{table}{Inference parameters for Kimi-K3 and DeepSeek-V4 Flash/Pro.}
    \label{tab:frontier-inference-settings}
\end{minipage}
\end{center}

\subsection{Expected Calibration Error}
For each generated sample $i$, let $c_i\in[0,1]$ denote the confidence predicted by the language model, and let $s_i\in[0,1]$ denote the normalized visual-quality score assigned by a vision evaluator. We partition the predictions into $M(M=10)$ equally spaced confidence bins $\{B_m\}_{m=1}^{M}$. For each non-empty bin $B_m$, the average confidence and average visual score are:
\begin{equation}
\small
\operatorname{Conf}(B_m)
=\frac{1}{|B_m|}\sum_{i\in B_m}c_i,
\qquad
\operatorname{Score}(B_m)
=\frac{1}{|B_m|}\sum_{i\in B_m}s_i.
\end{equation}

\noindent We define the Expected Calibration Error (ECE) as
\begin{equation}
\small
\operatorname{ECE}
=\sum_{m=1}^{M}
\frac{|B_m|}{N}
\left|
\operatorname{Score}(B_m)
-\operatorname{Conf}(B_m)
\right|,
\end{equation}
where $N$ is the total number of evaluation samples. A lower ECE indicates closer agreement between the model's self-reported confidence and the visual evaluator's assessment of generation quality.

\noindent For evaluators whose raw outputs do not naturally span $[0,1]$, we apply min--max normalization over the evaluation set before computing ECE:
\begin{equation}
\small
s_i
=\frac{r_i-\min_j r_j}
{\max_j r_j-\min_j r_j},
\end{equation}
where $r_i$ is the raw visual score. In all experiments, we use $M=10$ equal-width confidence bins.

\subsection{Remarks on PhysHack}

PhysHack refers to a systematic mismatch between \emph{checkable physical validity} and the broader objective of \emph{semantic and geometric correctness} in LEGO Brick Assembly. A generated program may satisfy local constraints, such as valid brick types, bounded coordinates, and collision-free placement, while still failing to preserve the object identity, major components, or global structure specified by the text prompt. Therefore, PhysHack should not be interpreted as a failure of the physical-validity evaluator itself. Rather, it arises because physical validity captures only one necessary component of successful LEGO generation and is insufficient as a standalone proxy for overall generation quality.

Our results provide complementary evidence for this phenomenon. First, models trained on the full dataset achieve high physical validity but substantially lower semantic and structural alignment. For example, the full-data Qwen model reaches $0.93$ Validity@4, while obtaining only $0.39$ Gemma@4 and $0.67$ DINOv3@4. Second, directly optimizing physical validity as the sole reinforcement-learning reward further amplifies this mismatch. In the physics-only setting, the model reaches $1.00$ Validity@4, but remains substantially weaker on Qwen-VL, DINOv3, and voxel alignment. These results demonstrate that improving a locally verifiable physical proxy does not necessarily improve the intended object-level outcome.

The data-selection experiments further suggest that PhysHack is partly induced by the composition of the training data. As illustrated in Figure~\ref{fig:origin_image}, the full training pool contains many trajectories that pass physical-validity checks but omit object-defining components or exhibit weak text--structure consistency. Consistent with this observation, models trained on low-value trajectories perform substantially worse in semantic alignment, whereas selecting semantically aligned and physically feasible trajectories allows a $5\%$ subset to outperform full-data training on multiple independent evaluators. We therefore use the term \emph{data-induced} to indicate that physically valid but semantically misaligned supervision is an important source of PhysHack, rather than to claim that dataset noise is its only possible cause.

More broadly, PhysHack can be understood as a domain-specific form of proxy optimization. In LBA, local physical constraints are easy to verify, while semantic fidelity and global geometry are harder to measure. A model can consequently learn shortcuts that maximize physical-validity scores without producing the intended structure. This failure mode can be learned from noisy supervision and further amplified when physical validity is optimized in isolation.

PVPO mitigates this behavior by coupling physical feasibility with voxel-space geometric alignment, thereby making the training signal more consistent with the complete generation objective. The improvements observed on held-out prompts, independent VLM evaluators, structural-stability metrics, and blinded human evaluation indicate that the resulting gains are not limited to the voxel reward or to the model used for data selection.

\subsection{Language Model Disclosure}
We use LLM to assist with minor manuscript polishing and LaTeX formatting. All technical content, experimental design, analysis, and final revisions were reviewed and verified by the authors.

\end{document}